\documentclass{article} 
\usepackage{tccml_iclr2025_conference,times}

\usepackage[linesnumbered,ruled]{algorithm2e}

\usepackage{amsmath,amsfonts,bm}

\def\eqref#1{equation~\ref{#1}}

\def\1{\bm{1}}

\DeclareMathAlphabet{\mathsfit}{\encodingdefault}{\sfdefault}{m}{sl}
\SetMathAlphabet{\mathsfit}{bold}{\encodingdefault}{\sfdefault}{bx}{n}

\usepackage{graphicx}
\usepackage{hyperref}
\usepackage{url}
\usepackage{enumitem}
\usepackage{subfigure}
\usepackage{tabu}
\usepackage{multirow}
\usepackage{multicol}
\usepackage{float}
\usepackage{makecell}
\usepackage{booktabs}
\usepackage{caption}
\title{Deep Reinforcement Learning for Power Grid Multi-Stage Cascading Failure Mitigation}

\author{Bo~Meng, Chenghao~Xu \& Yongli~Zhu\thanks{Corresponding author.} \\
School of System Science and Engineering\\
Sun Yat-sen University\\
Guangzhou, China\\
\texttt{\{mengb,xuchh29\}@mail2.sysu.edu.cn, yzhu16@alum.utk.edu}\\
}

\iclrfinalcopy 
\begin{document}

\maketitle

\begin{abstract}
Cascading failures in power grids can lead to grid collapse, causing severe disruptions to social operations and economic activities. In certain cases, multi-stage cascading failures can occur. However, existing cascading-failure-mitigation strategies are usually single-stage-based, overlooking the complexity of the multi-stage scenario. This paper treats the multi-stage cascading failure problem as a reinforcement learning task and develops a simulation environment. The reinforcement learning agent is then trained via the deterministic policy gradient algorithm to achieve continuous actions. Finally, the effectiveness of the proposed approach is validated on the IEEE 14-bus and IEEE 118-bus systems.
\end{abstract}

\section{Introduction}
\label{Introduction}
The modern large power grid consists of thousands of generators, substations, and transmission lines, all intricately interconnected and interdependent, working together to maintain the stable transmission of electricity. However, during the operation of the power system, various events may occur, among which \textit{cascading failures} is particularly complex and highly damaging \cite{Sun,Jyoti,Uwamahoro}. Cascading failures in power systems are typically triggered by the failure of a single component, e.g., a transmission line. These faults can rapidly \textit{propagate} through the tightly interconnected network, potentially causing severe disturbances across the entire power grid and even leading to a complete system collapse \cite{Mei,Zhang,Biwei}. Such kind of events can pose a significant threat to the security of the power grids and result in severe social and economic consequences.

Cascading failures can lead to devastating outcomes \cite{Hengdao,Salehpour}. For example, on June 19, 2024, at approximately 15:17, Ecuador experienced a nationwide blackout, resulting in a collapse of the nation's power grid, affecting around 18 million people, with the power outage lasting for approximately 3 hours. The direct cause of this incident was the failure of the Milagro-Zhoray transmission line, which triggered a series of cascading failures, ultimately resulting in a widespread outage. This severe outage underscores the importance of developing fast cascading failure mitigation strategies for complex power grids.

In the power system area, cascading failure mitigation refers to a series of control actions to prevent the chain reaction after the first fault (e.g., one-line tripping), thereby avoiding system-wide blackouts. In recent years, numerous studies have emerged in this field. For example, \cite{guo} proposed a method combining transient stability analysis with interaction graphs to identify critical lines and mitigate cascading failures by reducing the fault probability of components on these critical lines. \cite{li} applied network flow theory to study the process of power flow redistribution and proposed a cascading failure mitigation strategy based on adaptive power balance recovery and selective edge protection. Inspired by the propagation patterns of faults, \cite{Bhaila} employed graph neural networks (GNNs) to model and analyze cascading failures in power grids using an end-to-end approach. \cite{liu}, on the other hand, utilized an improved percolation theory to analyze the survivability of nodes in power grids and proposed an effective mitigation strategy.

In this paper, a deep reinforcement learning (DRL) approach is developed for mitigating multi-stage cascading failures (MSCF) in power systems, with the following contributions: (1) A simulation environment for multi-stage cascading failure study is constructed; (2) The Deep Deterministic Policy Gradient (DDPG) algorithm is adopted to address the MSCF issue; (3) The proposed model is validated on the IEEE 14-bus and 118-bus systems, demonstrating its effectiveness.

\section{Methodology}
\label{Methodology}

\subsection{Multi-Stage Cascading Failure (MSCF) Problem}

\begin{figure}
\centering
\includegraphics[width=1\linewidth]{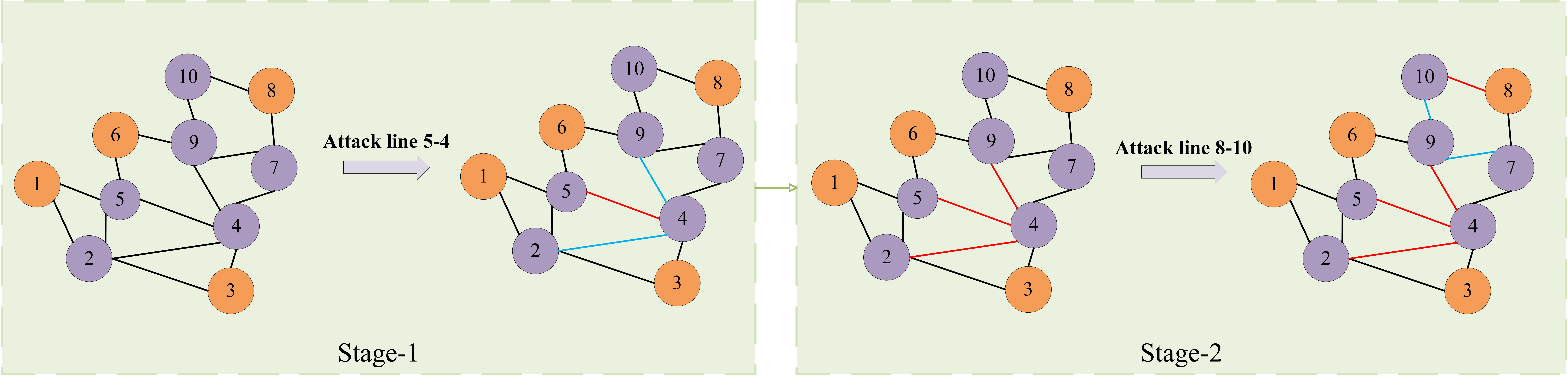}
\caption{\label{fig:multi-stage cascading failure}An example of a multi-stage cascading failure.}
\vspace{-10pt}
\end{figure}

Traditionally, single-stage cascading failure problems have been well studied \cite{Qi}. However, in certain situations, multiple stages may occur \cite{Zhu}. For example, Fig. \ref{fig:multi-stage cascading failure} depicts a multi-stage cascading failure example: an earthquake causes the loss of the power line 4-5, triggering the first stage of cascading failures (lines 2-4 and 4-9 are subsequently tripped due to the over-limit line power flow, \textit{after} the loss of line 4-5). Suppose the (remaining) power grid does not collapse and enters a steady state. Then, after a short period, the aftershock may break another line, triggering \textit {another stage} of cascading failures.  

One approach to handling the MSCF problem is to decompose them into multiple sub-problems of single-stage and then solve each by each. However, this way might overlook the interdependence between stages. On the other hand, if we map the concept of ``each \textit{stage}" to the concept of ``each \textit{step}" in the RL context, then the MSCF problem can be investigated \textit{holistically} under various mature frameworks of reinforcement learning, which is the motivation of this paper.

In this paper, the DDPG algorithm and the Actor-Critic framework are utilized \cite{ddpg_Timothy}, \cite{AC_Volodymyr}. The output of DDPG can be deterministic and real-number valued; hence, it performs well in solving problems with continuous actions \cite{DDPG}.

\subsection{Environment implementation}

In our work, a simulation environment is developed for MSCF mitigation using Python and Matpower, which is a well-known MATLAB toolbox for AC power flow (ACPF) computation. Cross-tool interaction and data communication between Python and MATLAB have been achieved via a Python-MATLAB handler. Several key designs regarding this environment are described below.

\subsubsection{Definitions of Step and Episode}
\textbf{Step}: a step means a \textbf{stage} when the power grid is attacked (e.g., by natural disasters), causing the grid to evolve into a new state (i.e., how many buses (i.e., nodes) and "lines" (i.e., edges) are still "available"; how many islands are formed; how large is the power flow on each remaining line; etc.).

\textbf{Episode}: an episode is one specific \textbf{set of steps} when the power grid is consecutively attacked. At the end of each episode, the final status is either ``Win" or ``Lose " (cf. definitions in later sections).

\subsubsection{State design}
For an \textit{n}-bus power grid, our \textit{state} is defined as follows:
\[state=[line\_status,P_1,Q_1,V_1,\theta_1,...,P_n,Q_n,V_n,\theta_n]\]
where, $line\_status$ is the percentage value obtained via dividing the actual line power flow by its maximum limit; $P_i,Q_i,V_i,\theta_i,(i=1,...,n)$ denotes the active power injection, reactive power injection, voltage magnitude and angle of the $i$-th bus, respectively.

\subsubsection{Action design}
Cascading failure might be mitigated by adjusting the generator's power generation. Thus, the generation coefficients $[a_1,…,a_m]$ of all \textit{m} generators are considered as the \textit{action}. The power output of the \textit{i}-th generator is the product of $a_i$ and its power capacity (i.e., the maximum power).

\subsubsection{Island detection and availability assessment}
The grid can become disconnected when lines are lost (due to an incident or line overload). Therefore, the first step is to assess the connectivity of the grid. To that end, we employ the union-find algorithm (c.f. Appendix \ref{Appendix:A1}) to locate all the remaining islands.

\begin{figure}
\centering
\includegraphics[width=1\linewidth]{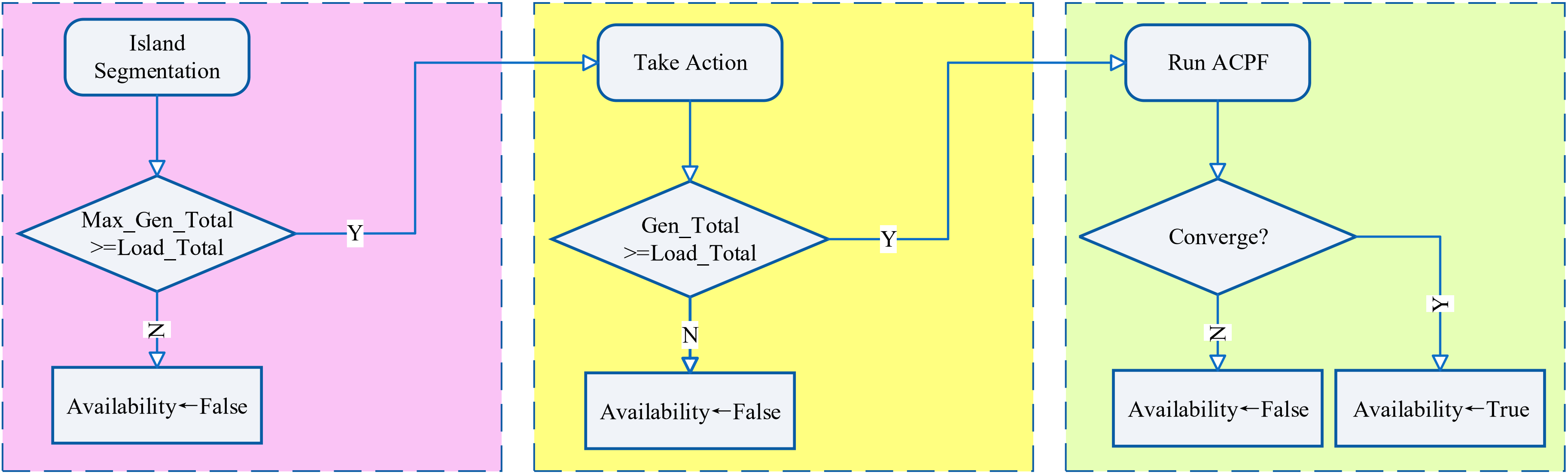}
\caption{\label{fig:Availability Assessment}Island Availability assessment.}
\vspace{-10pt}
\end{figure}

The ``availability" of an island means whether it is still \textit{alive} at the end of a specific cascading failure stage; if not, it will be discarded in later stages. The availability assessment is carried out after the island detection. The criteria for island availability are described in Fig. \ref{fig:Availability Assessment}. $Max\_Gen\_Total$ and $Gen\_Total$ are respectively the total power capacity and the total actual power output of all the remaining generators in a specific island, and $Load\_Total$ is the total load demand in that island.

\subsubsection{Reward design}
\begin{itemize}
\item \textbf{Total cost of generation: $-c_1\cdot cost$}.
Here, $c_1$ is a hyperparameter. $cost$ means the total generation cost (\$) of all islands whose \textit{availability} is true.
\item \textbf{Loss of load penalty: $-BaseReward_1\cdot P_{loss}/P_{total}$}.
$P_{loss}$ is the total load on \textit{unavailable} islands at current stage, while $P_{total}$ represents the original total load of initial power grid.
\item \textbf{Convergence reward: $BaseReward_2$}.
This reward is given when half or more of all the currently remaining islands have converged.
\item \textbf{Win reward: $BaseReward_3\cdot (P_{available}/P_{total})^{c_2}$}.
This reward is given when the win conditions are met. $P_{available}$ is the total load of \textit{available} islands.
\end{itemize}

\vspace{-3pt}

\begin{figure}
    \centering
    \subfigure[]{
    \begin{minipage}[t]{0.53\textwidth}
    \includegraphics[width=1\textwidth]{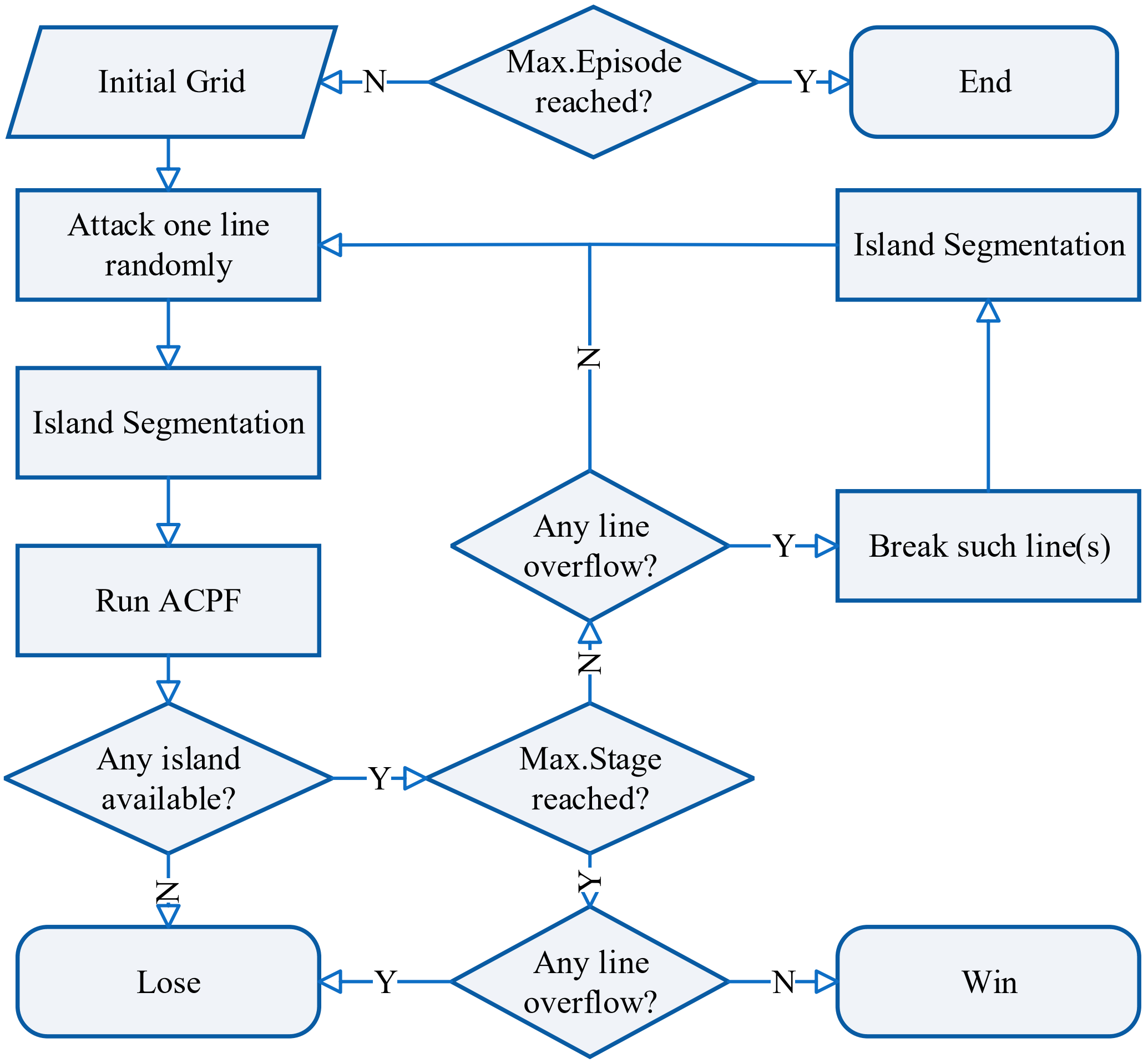}
    \end{minipage}
    \label{fig:workflow}
    }
    \subfigure[]{
    \begin{minipage}[t]{0.42\textwidth}
    \includegraphics[width=1\textwidth]{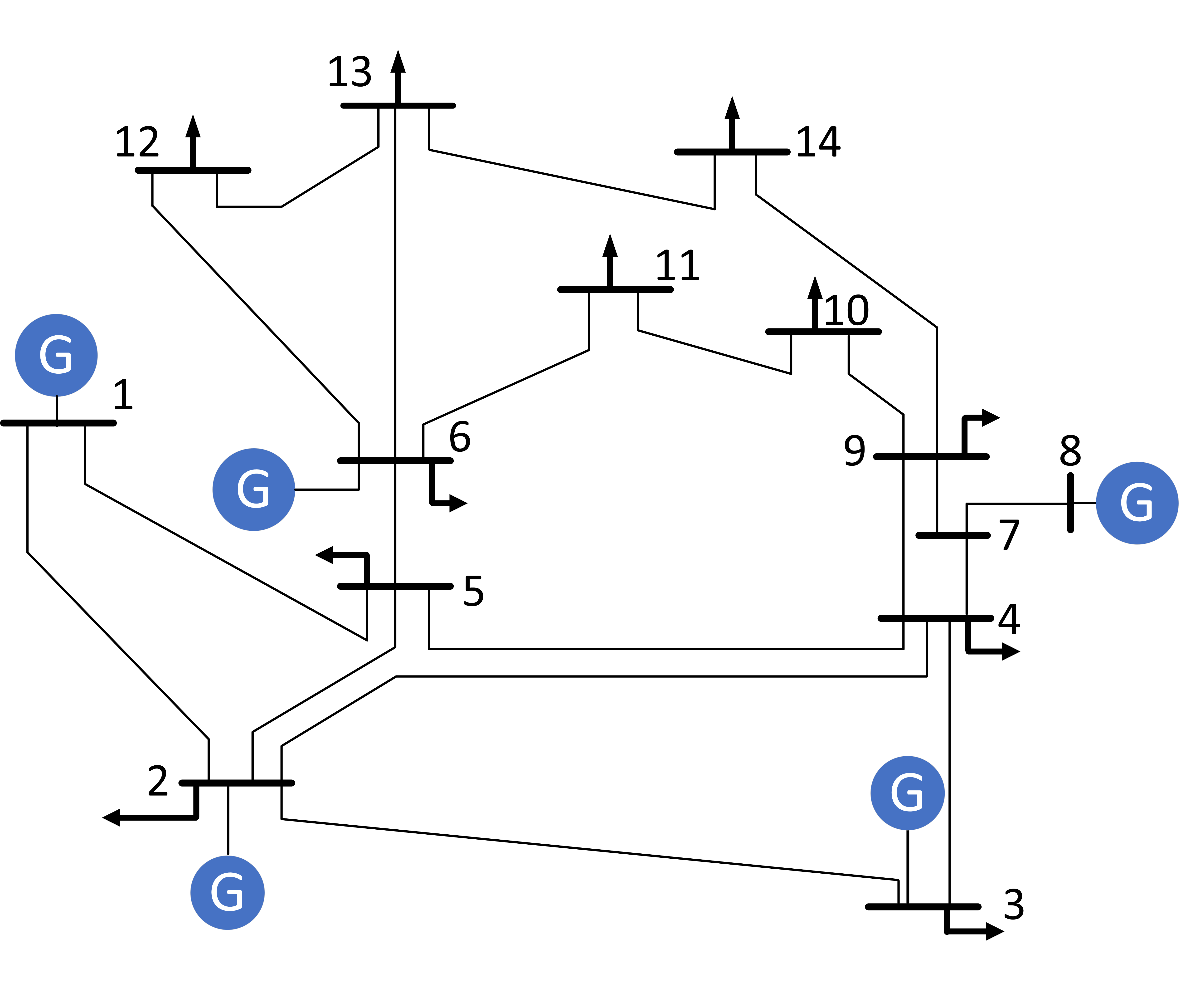}
    \end{minipage}
    \label{fig:14-bus}
    }
    \caption{(a) The overall workflow of grid simulation for MSCF study; (b) The IEEE 14-bus system.}
\end{figure}

Here, $c_1$, $c_2$, $BaseReward_1$, $BaseReward_2$, and $BaseReward_3$ are constants related to a specific power grid. A basic idea in picking those constants is to make the above four parts in the same order of magnitude. Finally, the overall workflow for the RL-based MSCF study is shown in Fig. \ref{fig:workflow}.

\section{Experiments and Results}
\label{Experiments}

The proposed approach is tested on the IEEE 14-bus and modified IEEE 118-bus systems. The IEEE 14-bus system has 5 generators and 20 lines, with its topology shown in Fig. \ref{fig:14-bus}. For other details about the experiment settings and hyperparameters, please refer to Appendix \ref{app:setup and parameters}.

For each power grid, a DRL model is trained for 300 episodes. After training, the model interacted with the environment for an additional 1000 episodes, during which the total reward in each episode is recorded, and the final win rate is computed.

\begin{figure}
    \centering
    \subfigure[]{
    \begin{minipage}[t]{0.47\textwidth}
    \includegraphics[width=1\textwidth]{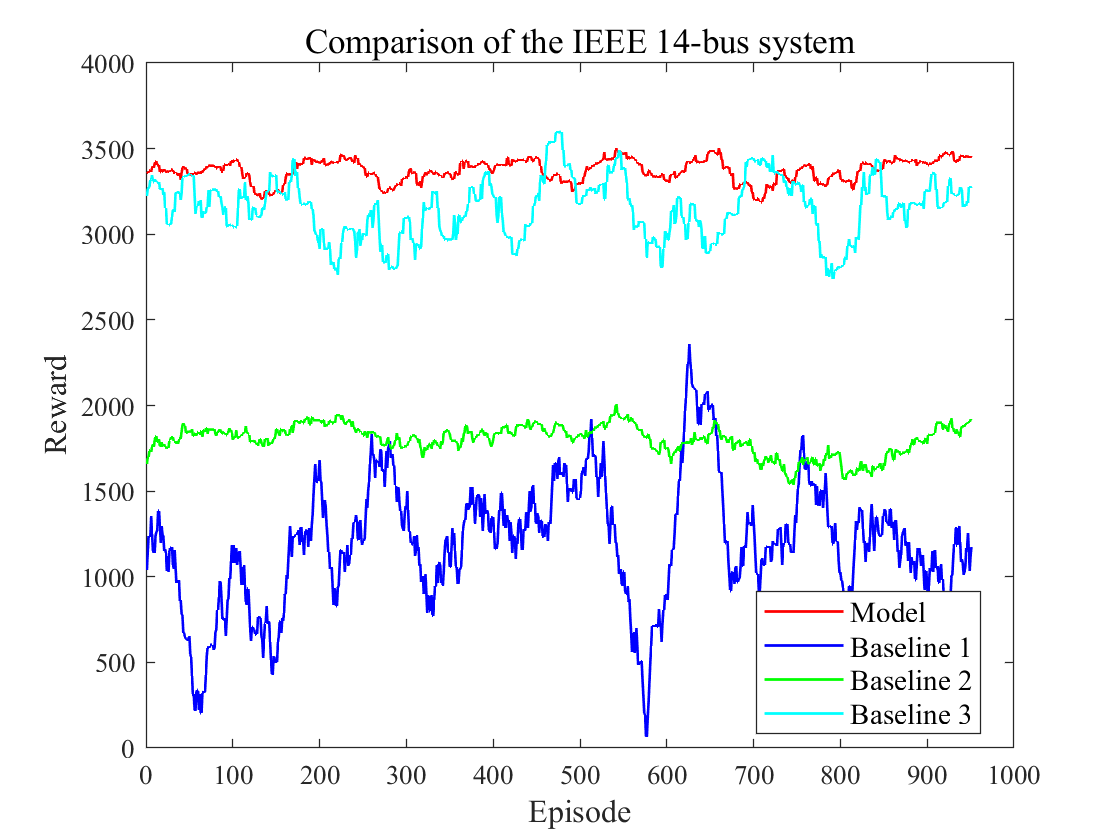}
    \end{minipage}
    \label{fig:14_result}
    }
    \subfigure[]{
    \begin{minipage}[t]{0.47\textwidth}
    \includegraphics[width=1\textwidth]{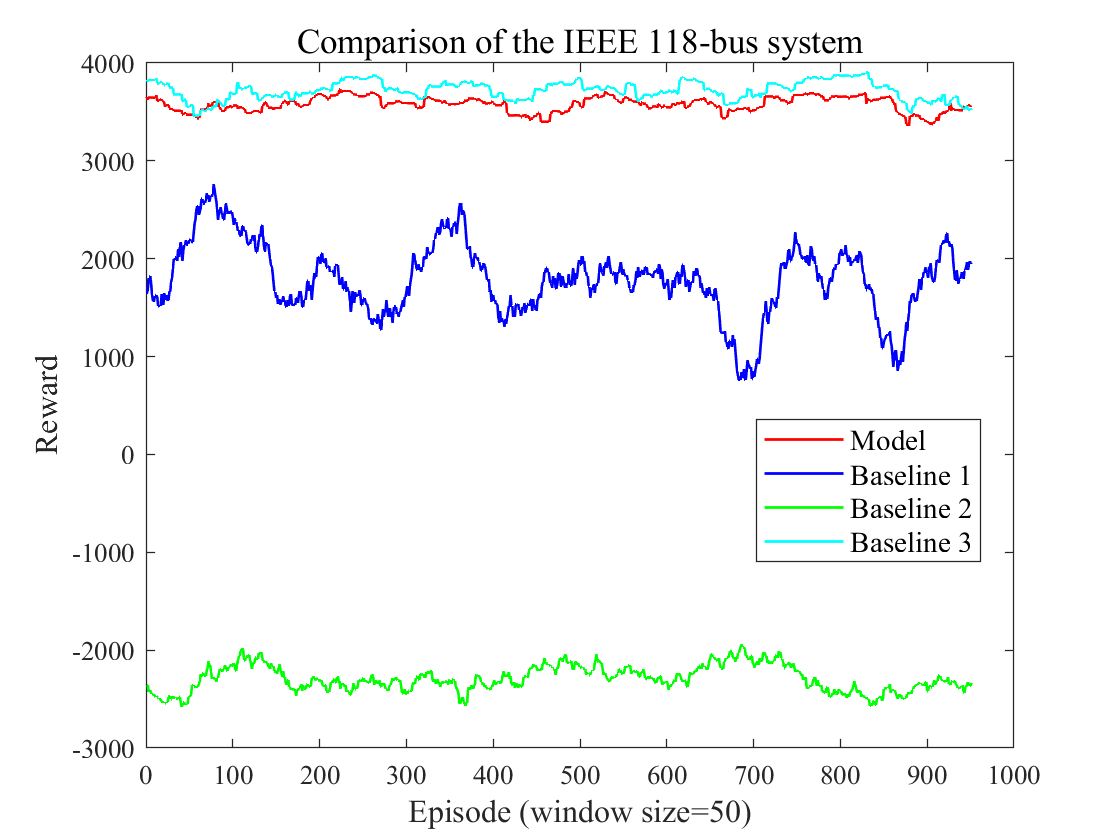}
    \end{minipage}
    \label{fig:118_result}
    }
    \vspace{-10pt}    
    \caption{The moving-average reward comparison.}
    \label{fig:result}
\end{figure}

The model is compared with three baseline strategies, as shown in Table \ref{tab:Win rate} and Fig. \ref{fig:result}. \textbf{Baseline 1} means each generator output a random power. \textbf{Baseline 2} means all generators output the maximum power. \textbf{Baseline 3} means all generators operate at half of their maximum power output. It can be observed that the DRL achieves a good performance, with the highest win rate, large average rewards, and more stable behaviors.
\begin{table}
\centering
\caption{\label{tab:Win rate}Win rate comparison.}
\begin{tabular}{c|c c}
\Xhline{2pt}
Method & IEEE 14-bus System &IEEE 118-bus System\\\hline
  \hline
DDPG & \textbf{95.5\%} & \textbf{97.8\%}\\
Baseline 1 & 52.0\% & 51.7\%\\
Baseline 2 & 93.3\% & 8.40\%\\
Baseline 3 & 85.6\% & 97.0\%\\
\Xhline{2pt}
\end{tabular}

\end{table}

\section{Conclusion}
\label{Conclusion}

This paper implements and validates a DRL-based solution for multi-stage cascading failure mitigation. One limitation of the current solution is that the differences in the states are relatively small, causing the majority of the model's actions to be similar. In future work, we will explore other state designs to improve the action's variability.

\bibliography{iclr2025_conference}
\bibliographystyle{iclr2025_conference}

\appendix
\section{Appendix}

\subsection{The Union-Find Algorithm for Power Grid Island Detection}\label{Appendix:A1}
The union-find algorithm is a data structure used to handle dynamic connectivity problems. Its basic idea is to determine whether elements belong to the same set recursively and to merge sets when necessary. Based on the results of island detection, the original grid may need to be divided into multiple islands, which provides the basis for later evaluation of the system status.

\begin{algorithm}
    \SetAlgoLined 
	\caption{Island Detection}
	\KwIn{A power grid $G$ with bus set $N$ and line set $E$}
	\KwOut{Islands $I$}
        Initialize an array $p$ such that $p[n]\gets n$ for all $n$\;
        \For{(u,v) in E}{
        Perform \textit{Union}$(u,v,p)$ to merge their sets\;
        }
        \For{n in N}{
        Perform \textit{Find}$(n)$ to determine the root\;
        }
        Group all buses by their root into disjoint sets $I$\;
\end{algorithm}

\subsection{The topology of the IEEE 118-bus system}
\begin{figure}
\centering
\includegraphics[width=0.94\linewidth]{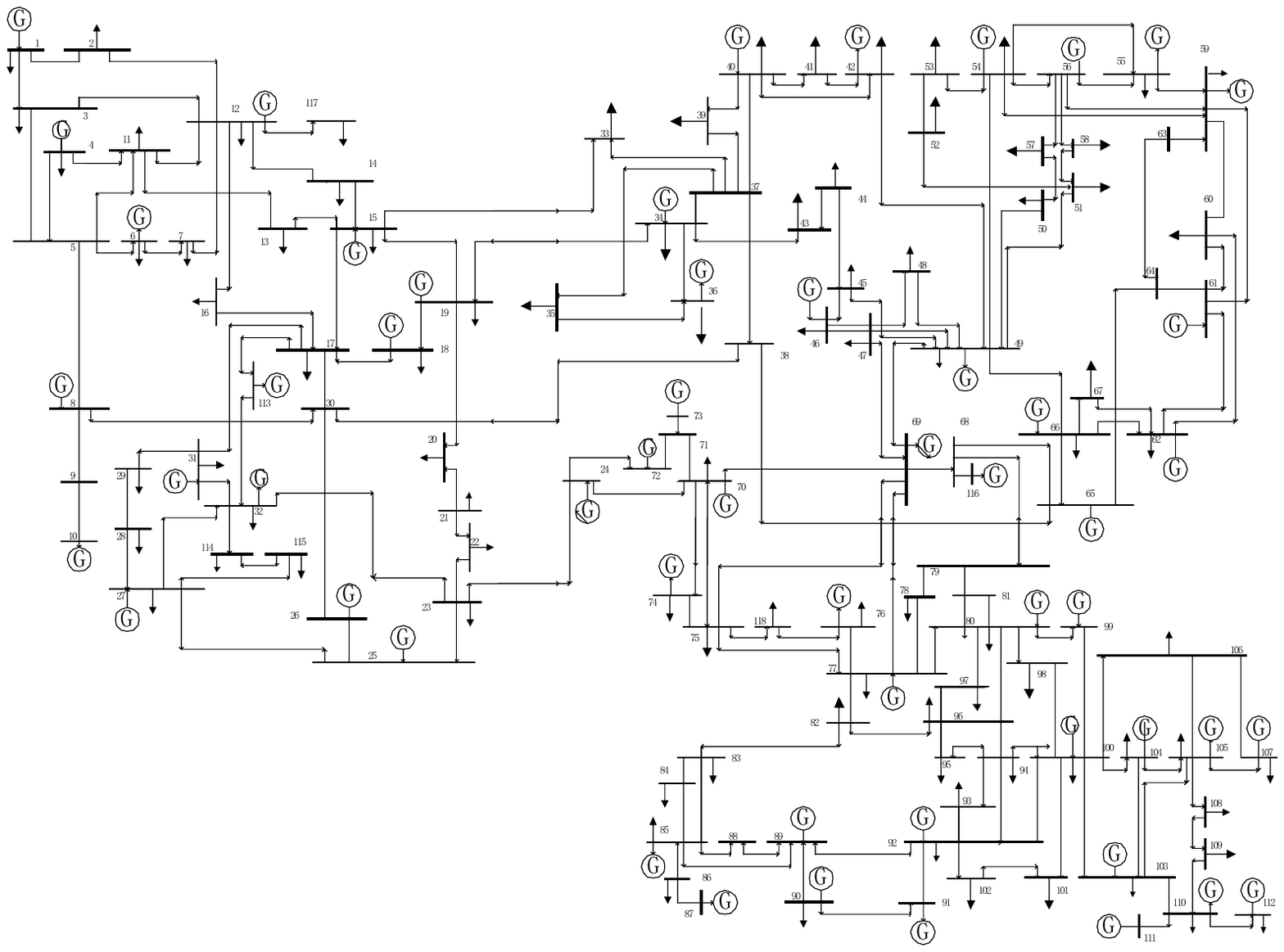}
\vspace{-10pt}
\caption{\label{fig:118-bus}The topology of the IEEE 118-bus system.}
\end{figure}

The topology of the IEEE 118-bus system is shown in Fig. \ref{fig:118-bus}.


\subsection{Reward comparison}
\begin{figure}
    \centering
    \subfigure[]{
    \begin{minipage}[t]{0.47\textwidth}
    \includegraphics[width=1\textwidth]{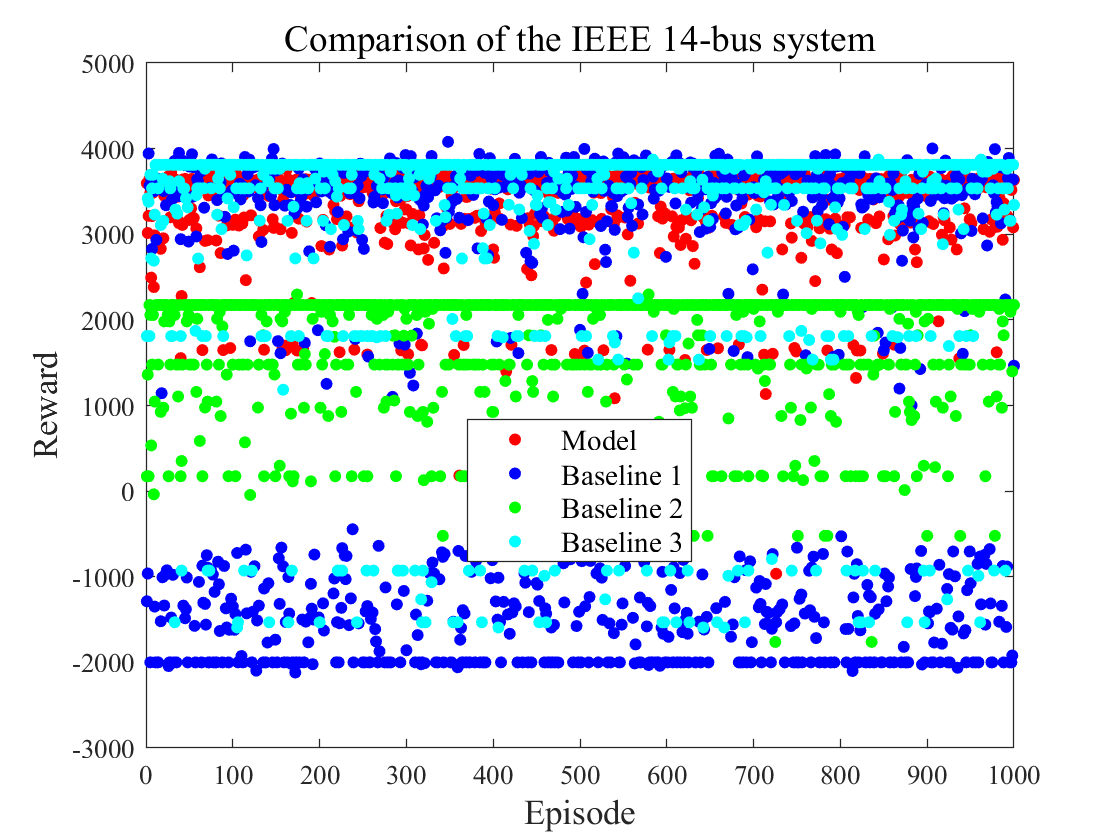}
    \end{minipage}
    \label{fig:14_result_point}
    }
    \subfigure[]{
    \begin{minipage}[t]{0.47\textwidth}
    \includegraphics[width=1\textwidth]{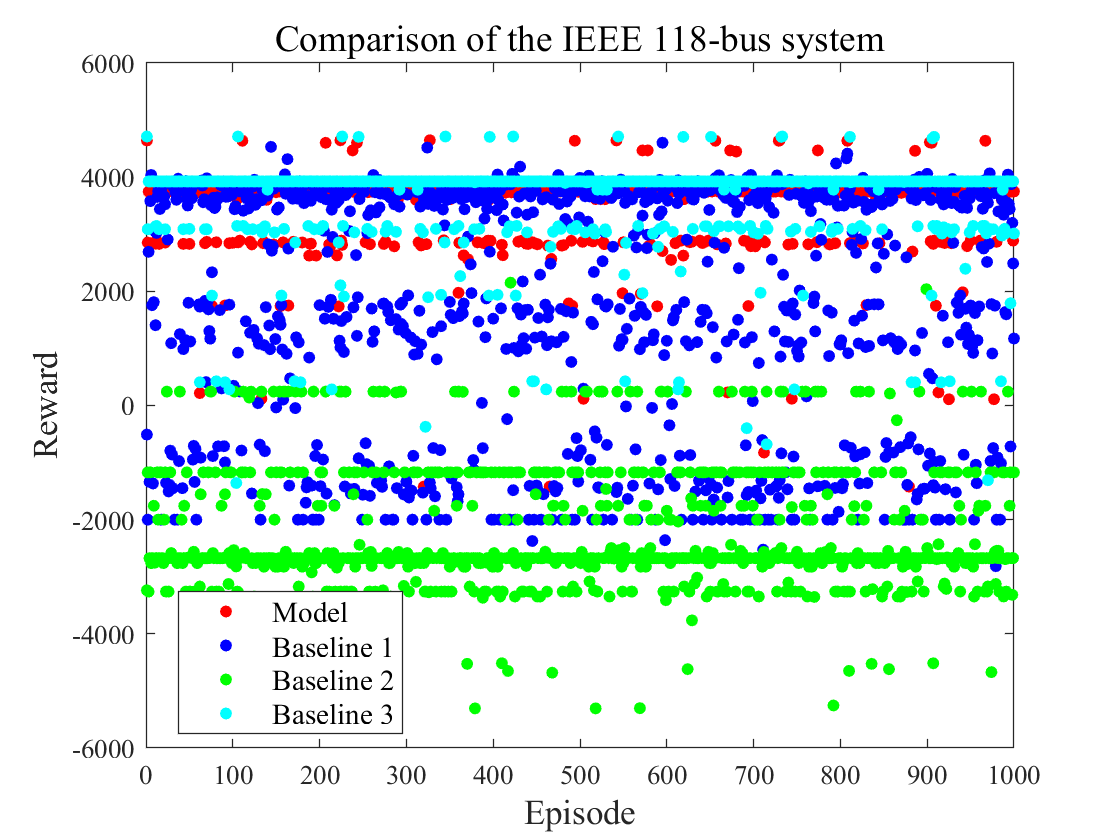}
    \end{minipage}
    \label{fig:118_result_point}
    }
    \vspace{-10pt}
    \caption{The reward comparison.}
    \label{fig:result_point}
\end{figure}

The reward comparison is shown in Fig. \ref{fig:result_point}.

\subsection{Cross-tool interaction}
\begin{figure}
\centering
\includegraphics[width=0.88\linewidth]{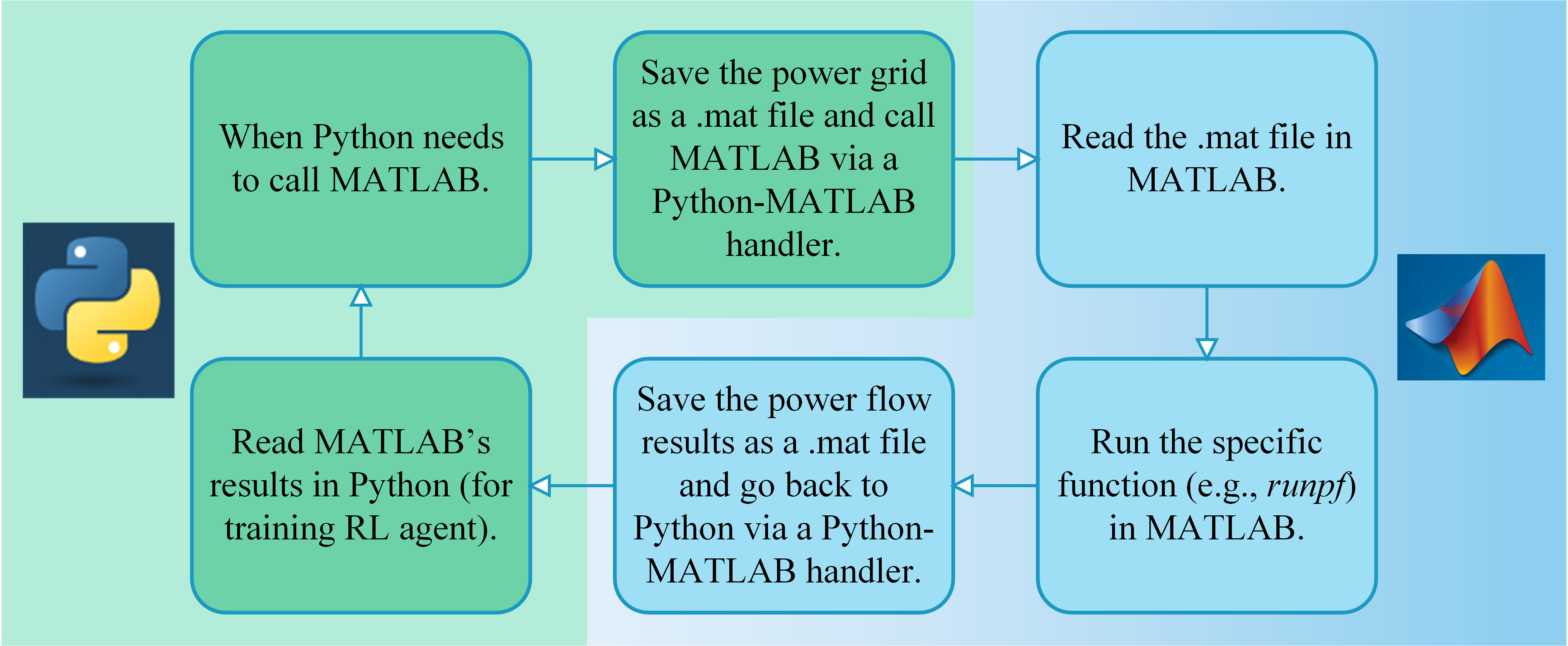}
\caption{\label{fig:interaction}The process of cross-tool interaction.}
\end{figure}

The process of cross-tool interaction is shown in Fig. \ref{fig:interaction}.

\subsection{Experiment Settings and Hyperparameters}
\label{app:setup and parameters}

The experiments are carried out on a computer with an Intel Core i5-12400F CPU, 32 GB RAM, and a GeForce RTX 4060ti GPU. The development environments are Python 3.11, PyTorch 2.3.1, and MATPOWER 8.0.

The IEEE 118-bus system contains 54 generators and 179 lines.  Its topology is shown in Fig. \ref{fig:118-bus}. The environment parameters for both 14-bus and 118-bus systems are summarized in Table \ref{tab:environment parameter}. $stage\_max$ represents the maximum number of stages in the MSCF problem, and $line\_limit$ refers to the maximum allowed power flow on the lines.

\vspace{8pt}

\begin{minipage}[c]{0.67\textwidth}
    \centering
    \captionof{table}{\label{tab:environment parameter}Environment parameters.}
\begin{tabular}{c|c c}
\Xhline{2pt}
Parameter & \makecell[c]{IEEE 14-bus\\System}  &\makecell[c]{IEEE 118-bus\\System} \\\hline
\hline
$stage\_max$ & 3 & 3\\
$line\_limit$ & 200 & 450\\
$c_1$ & 0.03 & 0.005\\
$c_2$ & 1.7 &1.7\\
$BaseReward_1$ & 2000 & 2000 \\
$BaseReward_2$ & 1000 & 1000 \\
$BaseReward_3$ & 2000 & 2000\\
\Xhline{2pt}
\end{tabular}

\end{minipage}
\begin{minipage}[c]{0.3\textwidth}
\centering
\captionof{table}{\label{tab:model parameter}Model parameters.}
\begin{tabular}{c|c}
\Xhline{2pt}
Parameter & Value\\\hline
\hline
learning rate & $1\times10^{-4}$ \\
batch size & 128\\
discount factor($\gamma$) & 0.99\\
update rate($\tau$) & 0.001\\
\Xhline{2pt}
\end{tabular}

\end{minipage}

\vspace{14pt}

The model is trained using the DDPG algorithm and the configured parameters are shown in Table \ref{tab:model parameter}. Depending on the complexity of a given power grid, the number of hidden-layer neurons can be adjusted and experimented for the best performance.

\end{document}